\documentclass[conference]{IEEEtran}
\IEEEoverridecommandlockouts
\usepackage{cite}
\usepackage{amsmath,amssymb,amsfonts}
\usepackage{algorithmic}
\usepackage{graphicx}
\usepackage{textcomp}
\usepackage{xcolor}
\usepackage{adjustbox}
\usepackage{multirow}
\usepackage{fancyhdr}
\usepackage{hyperref}

\def\BibTeX{{\rm B\kern-.05em{\sc i\kern-.025em b}\kern-.08em
    T\kern-.1667em\lower.7ex\hbox{E}\kern-.125emX}}
\begin{document}
% models?
\title{Controlling the Output of a Generative Model by Latent Feature Vector Shifting
% {\footnotesize \textsuperscript{*}Note: Sub-titles are not captured in Xplore and
% should not be used}
\thanks{The work presented was carried out in the framework of the Horizon Europe TERAIS project, under the Grant agreement number 101079338.\\
Permission to make digital or hard copies of all or part of this work for personal or classroom use is granted without fee provided that copies are not
made or distributed for profit or commercial advantage and that copies bear this notice and the full citation on the first page. Copyrights for components
of this work owned by others than the author(s) must be honored. Abstracting with credit is permitted. To copy otherwise, or republish, to post on
servers or to redistribute to lists, requires prior specific permission and/or a fee.\\
979-8-3503-4353-3/23/\$31.00 \copyright2023 IEEE\\
https://doi.org/10.1109/DISA59116.2023.10308936}
}

\author{\IEEEauthorblockN{Róbert Belanec, Peter Lacko, Kristína Malinovská}
\IEEEauthorblockA{\textit{Department of Applied Informatics} \\
\textit{Faculty of Mathematics, Physics and Informatics}\\
Bratislava, Slovakia \\
belanecrobert22@gmail.com, \{peter.lacko, kristina.rebrova\}@fmph.uniba.sk}
% \and
% \IEEEauthorblockN{3\textsuperscript{rd} Given Name Surname}
% \IEEEauthorblockA{\textit{dept. name of organization (of Aff.)} \\
% \textit{name of organization (of Aff.)}\\
% City, Country \\
% email address or ORCID}
% \and
% \IEEEauthorblockN{4\textsuperscript{th} Given Name Surname}
% \IEEEauthorblockA{\textit{dept. name of organization (of Aff.)} \\
% \textit{name of organization (of Aff.)}\\
% City, Country \\
% email address or ORCID}
% \and
% \IEEEauthorblockN{5\textsuperscript{th} Given Name Surname}
% \IEEEauthorblockA{\textit{dept. name of organization (of Aff.)} \\
% \textit{name of organization (of Aff.)}\\
% City, Country \\
% email address or ORCID}
% \and
% \IEEEauthorblockN{6\textsuperscript{th} Given Name Surname}
% \IEEEauthorblockA{\textit{dept. name of organization (of Aff.)} \\
% \textit{name of organization (of Aff.)}\\
% City, Country \\
% email address or ORCID}
}

\maketitle

\begin{abstract}
% This document is a model and instructions for \LaTeX.
% This and the IEEEtran.cls file define the components of your paper [title, text, heads, etc.]. *CRITICAL: Do Not Use Symbols, Special Characters, Footnotes, 
% or Math in Paper Title or Abstract.
%State of the art generative models as StyleGAN3 \cite{karras2021alias} use latent feature vectors as basis for fotorealistic image generation. However, the ability to control the output is limited. Our goal was to design a methot for latent vector shifting for cotrolled output image modification.

% State-of-the-art generative models (e.g. StyleGAN3 \cite{karras2021alias}) often generate photorealistic images based on vectors sampled from their latent space.

State-of-the-art generative models (e.g. StyleGAN3 \cite{karras2021alias}) often generate photorealistic images based on vectors sampled from their latent space. However, the ability to control the output is limited. Here we present our novel method for latent vector shifting for controlled output image modification utilizing semantic features of the generated images. In our approach we use a pre-trained model of StyleGAN3 that generates images of realistic human faces in relatively high resolution. We complement the generative model with a convolutional neural network classifier, namely ResNet34, trained to classify the generated images with binary facial features from the CelebA dataset. Our latent feature shifter is a neural network model with a task to shift the latent vectors of a generative model into a specified feature direction. We have trained latent feature shifter for multiple facial features, and outperformed our baseline method in the number of generated images with the desired feature. To train our latent feature shifter neural network, we have designed a dataset of pairs of latent vectors with and without a certain feature. Based on the evaluation, we conclude that our latent feature shifter approach was successful in the controlled generation of the StyleGAN3 generator.
\end{abstract}

\begin{IEEEkeywords}
generative adversarial networks, controlled generation
\end{IEEEkeywords}

\section{Introduction \label{sec:intro}}
People often use drawing to represent and share information. Looking back, we can see that our predecessors often used drawing during rituals to interpret information that was probably meaningful for them and better than spoken word \cite{brumm2021oldest, whitley2016introduction}. In the modern world, there are ways to represent information, and visual representation is one of them. Visual information is expressed in digital format (e.g., images, websites) or physical format (e.g., paintings, sketches, posters). The current state of technology allows us to create software capable of transforming written information into a digital image that retains this information \cite{rombach2022high}.

In the past decade, with the increasing performance of GPUs, we have witnessed a significant advancement in generative models mainly because of the invention of Generative Adversarial Networks (GANs) \cite{goodfellow2014generative}. The task of this generative model is to learn the mapping from a specific latent (hidden) space to a meaningful image space based on a given image distribution. We wanted to take advantage of high-fidelity images generated by the generative model.

Most common GAN model architectures use two neural networks: a generator and a discriminator. The task of the generator is to generate an image that is impossible to distinguish as not being a real image from the dataset. The discriminator has the task of determining whether the generator generated the image by comparing it with the real image from the dataset. Both the generator and the discriminator are performing unsupervised learning, where the generator is trying to fool the discriminator, and the discriminator is trying to outsmart the generator. In other words, the generator learns how to produce realistic images from random noise vectors by playing this adversarial min-max game \cite{goodfellow2014generative}.

The primary GAN architecture has gone through a series of improvements. In the year of writing this paper, we can see GANs that can generate images almost indistinguishable from camera images. Among current image generating, state-of-the-art models based on GAN architecture are StyleGAN \cite{karras2021alias}, StyleSwin \cite{zhang2022styleswin} and ProjectedGAN \cite{sauer2021projected}.

The aim of our work is to drive the generator's latent representations in the desired direction to control the content of the generated output by selecting semantic features that we want to appear on the generated image. This can be implemented as either conditional or controlled generation. To create images with custom facial characteristics but still maintain the high fidelity of generated images, we propose a novel framework based on the StyleGAN3 generator for searching and editing semantic features of generated images. 

\section{Related work}
\subsection{Generative adversarial networks}
Most current state-of-the-art GAN designs focus primarily on increasing the fidelity of generated images \cite{brock2018large, 10.1145/3528233.3530738}, rather than controlling the output's content-wise. In the paper \cite{karras2021alias}, the authors of StyleGAN improved their StyleGAN2 \cite{karras2020analyzing} architecture, introducing StyleGAN3 to be. GAN architectures have also been used for other applications, such as image editing or video generation \cite{zhu2020domain, cheng2020sequential}.

\subsection{Image classification}
There are many \cite{yu2022coca, szegedy2017inception} models that are trained to generally classify objects, e.g., on Imagenet dataset \cite{imagenet_cvpr09}. However, some of the best-performing models are either transformer or encoder-decoder-based models capable of tasks beyond simple image classification \cite{liu2022swin}. It is also worth mentioning that better classification performance was recently achieved by developing new optimization algorithms like the Lion \cite{chen2023symbolic}. 
% toto uz nie je len klasifikacia, to je praveze blizky pribuzny alebo predchodca nasho pristupu
%In paper \cite{shen2020interpreting}, authors used a variation of Residual Network (ResNet) convolutional classifier \cite{6795724, he2016deep}. 

\subsection{Semantic feature control}
Shen and colleagues \cite{shen2020interpreting} proposed a framework to find and control the semantic features in the latent space of a generative model. They proposed and empirically corroborated a hypothesis that a hyperplane exists in the latent space that serves as a separation boundary for any binary semantic attribute. To find these boundaries, they have trained a set of support vector machines (SVM) \cite{boser1992training}, each for one semantic attribute. The inputs to the SVM were pairs of noise vectors $z$ and attributes predicted from the ResNet50 convolutional classifier. After finding the separation hyperplane, they modified the image by shifting the original latent vector $z$ in the direction of the separation hyperplane, thus making the binary attribute to emerge in the generated image. The cavity of this approach is, that the latent space of the generative model is often more complex, and the binary attributes are entangled, so that separation by a hyperplane is insufficient; therefore, the generated image modified by this strategy may contain unwanted attributes.

The latent space of generative models is often treated as a Riemannian  manifold \cite{arvanitidis2018latent, shao2018riemannian}. Some of the latest works focus mainly on extracting features from the latent space by supervised methods or finding a nonlinear relationship between different parts of the latent space and generated outputs \cite{voynov2020unsupervised, mukherjee2019clustergan, weder2021neuralfusion}. Currently, several methods \cite{bau2019gandissect, shen2020interpreting, broad2021network} have been found for controlling the generative process, but the principle of image synthesis is still not fully known or understood. This also suggests that separating binary attributes in latent space using hyperplanes may be insufficient and leads us to look for a more complex and non-linear solution.

\section{Shifted latent vectors dataset\label{sec:dataset}}
To be able to train a model of neural network that would shift the latent vectors in the desired direction, we have decided to build a new dataset using linear regression to find the feature axis vector, which represents the direction of a certain feature. In the latent space, this direction is probably not linear. Therefore, our approach using linear regression was unsuccessful. % for every image. 
Subsequently, we have decided to use the feature axis regression alongside the ResNet34 classifier to create a dataset for our latent feature shifter neural network model. In this section, we will walk through the generation of a newly shifted latent vectors dataset, which will contain four variations for each latent vector pair (vector without the feature, vector with the feature).

As mentioned in section \ref{sec:intro}, the GAN generator generates images from vector $Z$. Here $Z \subseteq \mathbb{R}^d$ stands for the GAN generating images from $d$-dimensional space, generated from Gaussian distribution. StyleGAN3 operates with a 512-dimensional space; therefore $d = 512$. We can also represent the generator as a function $g: Z \rightarrow X$ that maps the 512-dimensional latent space into a certain image space $X$, where each sample $x$ has certain semantic features (i.e. eyeglasses, smile, beard, etc.). We also have a scoring function $f_s: X \rightarrow S$, where $S \subseteq R^m$ represents our semantic feature space with $m$ features. We can therefore determine the magnitude of the presence of a particular semantic feature of a vector $Z$ by simply applying $f_s(g(Z))$.

\subsection{Generating unconditioned facial images using StyleGAN3}
To generate high-quality images of human faces, we have decided to use StyleGAN3 as it provides high fidelity and diverse results. Our previous successful approach was based on StyleGAN2, and other studies on latent space exploration also use StyleGAN models. Following this line of research we have decided to continue using StyleGAN3 instead of the current state-of-the-art GAN model StyleGAN-XL \cite{10.1145/3528233.3530738}.

Zhu an colleagues presented the StyleGAN3 model including with their implementations in the PyTorch framework\footnote{https://github.com/NVlabs/stylegan3} along with examples of image generation and pre-trained model weights. Due to a lack of time and computing resources, we have decided to use the pre-trained models provided by the authors. From the provided models trained on different datasets, but we chose the StyleGAN3 model with the configuration trained on the unaligned FFHQ dataset\footnote{https://github.com/NVlabs/ffhq-dataset} with the capability to generate images with a resolution of 1024x1024 marked as \textit{stylegan3-r-ffhqu-1024x1024}. As we have noticed, many facial artifacts emerge when using default truncation PSI (1.0). Therefore, we have chosen to change it to 0.7, which will produce less diverse images but with higher quality.

\subsection{Classifying generated images \label{sec:des_classifier}}
To classify the generated images, we used the ResNet34 classifier as ResNet classifier variations were used previously to extract features for SVM training \cite{shen2020interpreting}. Our ResNet34 classifier was trained using transfer learning (pre-trained on the ImageNet dataset \cite{imagenet_cvpr09}) by replacing the fully connected output layer with a dropout layer with dropout of 0.2 and a fully connected linear output layer with ten neurons. As we would like to teach the network to predict ten binary attributes, we have added a sigmoid activation function after the last layer. The network takes as the input an image and produces an output of ten numbers, each representing a probability of a certain image having a certain feature.

\subsection{Feature axis regression \label{sec:feature_axis}}
In this section, we describe the feature axis regression method we used to create the shifted latent vectors dataset. This approach was established by Shen et al. \cite{shen2020interpreting} who used a SVM regressor rather than linear regression. Roughly said, in our approach, we form a vector pointing in the direction of the feature we fit the line through when using linear regression. 

The relationship between input latent vectors $Z={z_1,z_2,...,z_n}$ and label vector from ResNet34 classifier, denoted as $Y={y_1,y_2,...,y_n}$, can be represented using a mathematical function in the form of a linear regression model, $Y=Z\beta+\epsilon$, where $\beta$ is a dimensional parameter vector consisting of coefficients for each dimension and $\epsilon$ denoting the error term (or intercept term). The linear regression model is fitted by estimating the regression coefficients $\beta$ that minimize the error $\epsilon=Y-Z\beta$. The transposed $\beta$ obtained from the fitted regression represents our feature axis. Subsequently, we normalize the feature axis because we will be using it to transform feature vectors, and we would like only to change their direction and not their size. After obtaining the feature axis, we modify the original latent vector by adding the feature axis to it as $Z_{new} = Z + \beta n$.

\subsection{Dataset compostion}
\begin{figure}
    \begin{centering}
        \includegraphics[width=0.9\columnwidth]{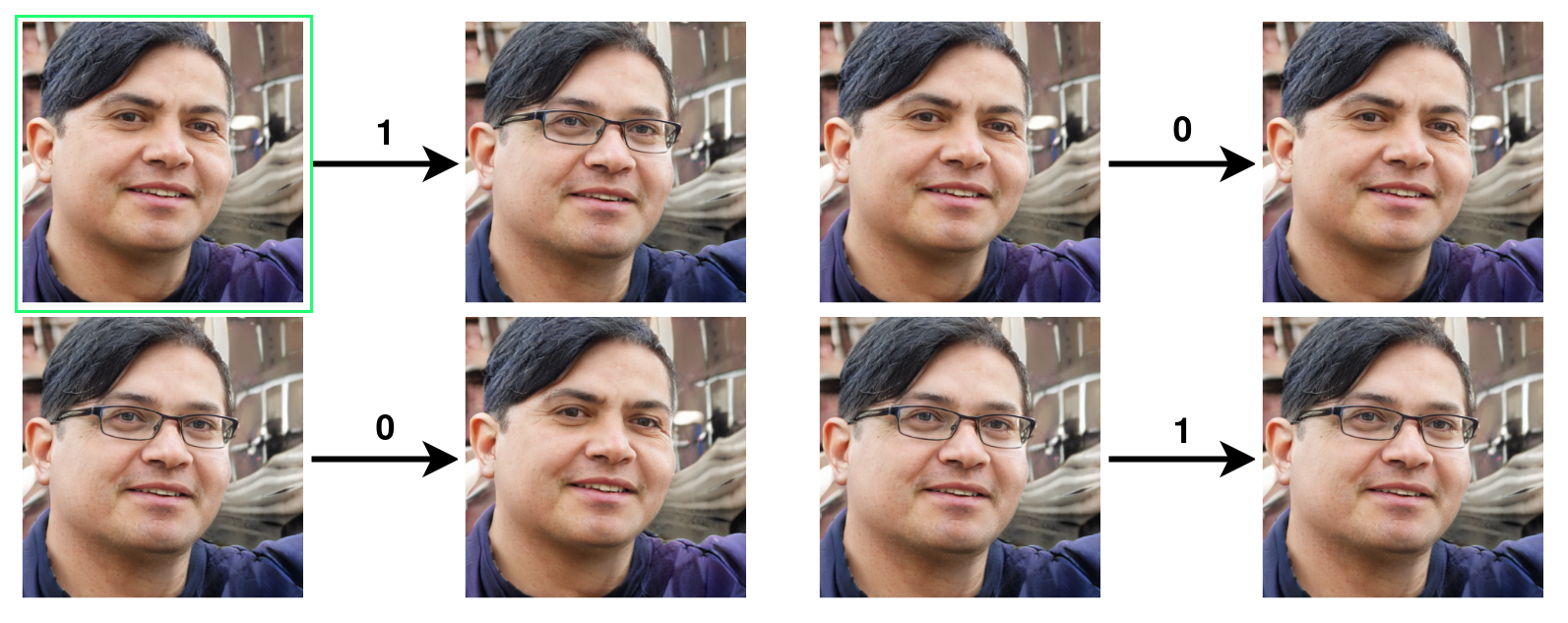}
        \caption{Example of generating four pairs samples of shifted latent vectors dataset from the original image (with green border).\label{fig:mld_pairs}}
    \end{centering}
\end{figure}

As a first step of generating our dataset, we have only selected latent vectors which generated images that did not have the selected feature (the classifier predicted the feature with a lower number than a threshold). For this purpose, we have set the threshold to be 0.5. After that, we used our feature axis regression approach to generate pairs (with and without the feature) for three selected features, namely the eyeglasses, males, and black hair. We have also transformed the features from classifier $Y$ by using the arctanh function, which should serve as an amplifier, before fitting the linear regression. After that, we generated the images for the shifted latent vectors and selected only the shifted latent vector classified by the ResNet34 classifier with a number greater than the threshold (the threshold is still 0.5). We have also selected latent vectors representing the shifted images without the feature (or not shifted to the feature axis direction). Subsequently, we have collected the pairs of a latent vectors without the desired feature and the same latent vectors shifted to the desired feature direction with feature axis regression method. Next, from each pair, we will generate four pairs of latent vectors (see the example in Figure \ref{fig:mld_pairs}) by combining latent vectors with and without the certain feature and with a binary label that represents whether the latent vector had the desired feature (1 means that we want a feature vector with the desired feature on the model output, 0 means the opposite) or that the feature vector did not manifest the desired feature on the model output.

\section{Our latent feature shifting approach}
\begin{figure*}
    \begin{centering}
        \includegraphics[width=0.7\textwidth]{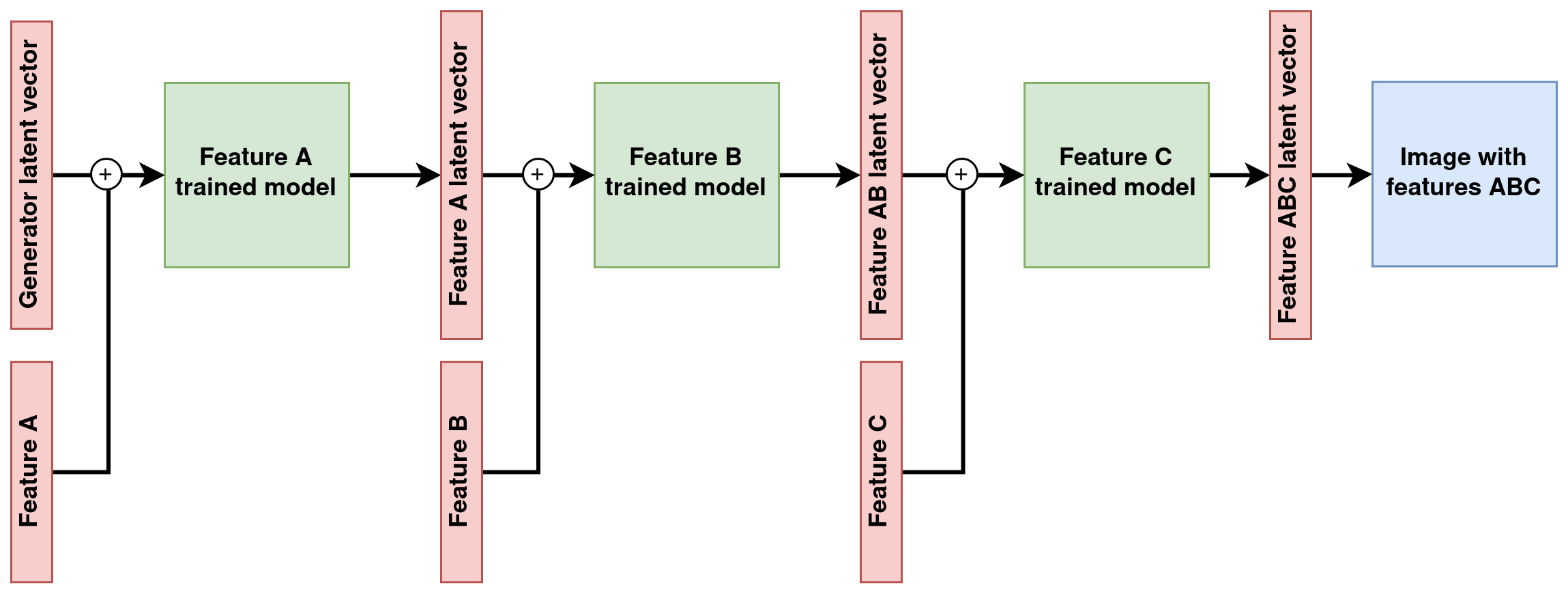}
        \caption{Diagram representing the process of generating a latent vector that will be shifted by each model trained on a different feature dataset, which should result in a latent vector representing all of the required features. The plus sign represent a vector concatenation operation.\label{fig:mlv_multiple}}
    \end{centering}
\end{figure*}

Since the relationships between the features in the latent space are probably not linear, we have chosen to use a neural network to move the latent vectors to contain the right feature. We have decided to use a simple method to train the neural network on the shifted latent vectors dataset described in section \ref{sec:dataset}. We are purposefully using a singular form of a feature, as we have designed the dataset to contain only images with a single feature, and created datasets for each feature separately. In the future, we can overcome this by training multiple models, each on a different feature, and joining them one after another so that the final vector would contain all of the features similar to Figure \ref{fig:mlv_multiple}.

We have created three shifted latent vectors datasets for the selected features: eyeglasses, male and black hair. Details about this dataset are shown in Table \ref{tab:mld}\footnote{Our datasets can be downloaded from \url{http://data.belanec.eu/mld.tar.gz}}.

\begin{table}
    \caption{Details about our shifted vectors dataset.}
    \label{tab:mld}
    \begin{adjustbox}{width=\columnwidth, center}
        \begin{tabular}{lccc}
        \multicolumn{4}{c}{\textbf{Shifted latent vectors dataset}}                   \\ \hline
        Feature                                  & Eyeglasses & Male   & Black hair \\
        Total tuples (shifted, feature, not shifted) & 92995      & 78405  & 82472      \\
        Total samples                            & 371980     & 313620 & 329888     \\
        Serialized object size in GB             & 2.9        & 2.5    & 2.6        \\ \hline
        \end{tabular}
    \end{adjustbox}
\end{table}

\subsection{Latent feature shifter design and training}
\begin{figure}[htb]
    \begin{centering}
        \includegraphics[width=\columnwidth]{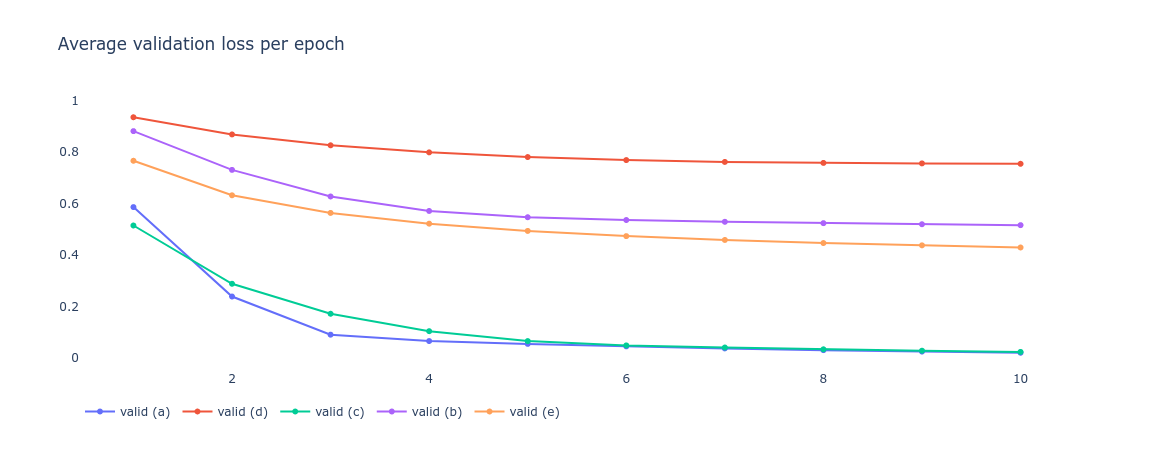}
        \caption{Average validation MSE loss development over ten training epochs for five different shifting latent vector model architectures (a-e).\label{fig:mlv_valid_loss}}
    \end{centering}
\end{figure}

We have trained five different neural network models (a to e) on shifted latent vectors dataset for eyeglasses feature with a random split of data into training, validation, and test subsets by ratio 80\%:10\%:10\% and mini-batches of size 16. Each model architecture can be seen in Table \ref{tab:mlnn_architectures}. Training took ten epochs using Adam optimizer with a learning rate of 0.00001 and MSE loss function. Test results are displayed in Table \ref{tab:moved_metrics}, and average validation MSE losses for each model are displayed in Figure \ref{fig:mlv_valid_loss}. Models \textit{a} and \textit{c} had similar test results. However, we will still prefer the model \textit{a}, as it has less trainable parameters, which means it can move the latent vectors faster. After finding the best model architecture, we decided to train the best architecture on a dataset with shifted latent vectors representing different features. We have trained architecture \textit{a} to move eyeglasses, male, and black hair features. Test results for architecture \textit{a} are shown in Table \ref{tab:moved_datasets_results}.

\begin{table*}
    \caption{Architectures of our trained shifting latent neural network models (a-e).}
    \label{tab:mlnn_architectures}
    \begin{adjustbox}{width=0.7\textwidth, center}
        \begin{tabular}{|ccccc|}
        \hline
        \multicolumn{5}{|c|}{\textbf{Shifting latent vectors neural network architectures}}                                                                                                                                 \\ \hline
        \multicolumn{1}{|c|}{a}                       & \multicolumn{1}{c|}{b}                       & \multicolumn{1}{c|}{c}                       & \multicolumn{1}{c|}{d}                       & e                    \\ \hline
        \multicolumn{1}{|c|}{FC 512 + k}              & \multicolumn{1}{c|}{FC 512 + k}              & \multicolumn{1}{c|}{FC 512 + k}              & \multicolumn{1}{c|}{FC 512 + k}              & FC 512 + k           \\ \hline
        \multicolumn{1}{|c|}{FC 1024 (ReLU)}          & \multicolumn{1}{c|}{FC 256 (ReLU)}           & \multicolumn{1}{c|}{FC 1024 (ReLU)}          & \multicolumn{1}{c|}{FC 256 (Leaky ReLU)}     & FC 1024 (Leaky ReLU) \\ \hline
        \multicolumn{1}{|c|}{\multirow{6}{*}{FC 512}} & \multicolumn{1}{c|}{\multirow{6}{*}{FC 512}} & \multicolumn{1}{c|}{FC 2048 (ReLU)}          & \multicolumn{1}{c|}{FC 128 (Leaky ReLU)}     & Dropout 0.2          \\ \cline{3-5} 
        \multicolumn{1}{|c|}{}                        & \multicolumn{1}{c|}{}                        & \multicolumn{1}{c|}{FC 1024 (ReLU)}          & \multicolumn{1}{c|}{FC 256 (Leaky ReLU)}     & FC 1024 (Leaky ReLU) \\ \cline{3-5} 
        \multicolumn{1}{|c|}{}                        & \multicolumn{1}{c|}{}                        & \multicolumn{1}{c|}{\multirow{4}{*}{FC 512}} & \multicolumn{1}{c|}{\multirow{4}{*}{FC 512}} & Dropout 0.2          \\ \cline{5-5} 
        \multicolumn{1}{|c|}{}                        & \multicolumn{1}{c|}{}                        & \multicolumn{1}{c|}{}                        & \multicolumn{1}{c|}{}                        & FC 1024 (Leaky ReLU) \\ \cline{5-5} 
        \multicolumn{1}{|c|}{}                        & \multicolumn{1}{c|}{}                        & \multicolumn{1}{c|}{}                        & \multicolumn{1}{c|}{}                        & Dropout 0.2          \\ \cline{5-5} 
        \multicolumn{1}{|c|}{}                        & \multicolumn{1}{c|}{}                        & \multicolumn{1}{c|}{}                        & \multicolumn{1}{c|}{}                        & FC 512               \\ \hline
        \end{tabular}
    \end{adjustbox}
\end{table*}

\begin{table}
    \caption{Test results of our trained latent feature shifter models. We have measured MSE as loss, MAE metric, and $R^2$ metric. The notable number is also the number of learnable parameters of the model.}
    \label{tab:moved_metrics}
    \begin{adjustbox}{width=0.8\columnwidth, center}
        \begin{tabular}{llllll}
        \multicolumn{6}{c}{\textbf{Shifting latent vectors model - eyeglasses}}         \\ \hline
        Architecture         & \textbf{a}       & b      & c       & d      & e       \\ \hline
        MSE loss             & 0.021   & 0.517  & 0.025   & 0.755  & 0.430   \\
        MAE                  & 0.091   & 0.573  & 0.102   & 0.693  & 0.523   \\
        $R^2$                & 0.978   & 0.465  & 0.974   & 0.225  & 0.560   \\
        Parameters           & 1051136 & 263168 & 5248512 & 329088 & 3150336 \\ \hline
        \end{tabular}
    \end{adjustbox}
\end{table}

\begin{table}
    \caption{Test results of our trained shifting latent vector model (\textit{a} architecture) on three different datasets (described by Table \ref{tab:mld}). We have measured MSE as loss, MAE metric, and $R^2$ metric.}
    \label{tab:moved_datasets_results}
    \begin{adjustbox}{width=0.6\columnwidth, center}
        \begin{tabular}{lccc}
        \multicolumn{4}{c}{\textbf{Shifting latent vectors model a}} \\ \hline
        Dataset      & Eyeglasses     & Male      & Black hair     \\ \hline
        MSE loss     & 0.021          & 0.067     & 0.035          \\
        MAE          & 0.090          & 0.204     & 0.126          \\
        $R^2$        & 0.978          & 0.838     & 0.964          \\ \hline
        \end{tabular}
    \end{adjustbox}
\end{table}

\section{Preliminary evaluation \label{sec:eva_manual}}
In this section, we evaluate the results of our approach by visually and manually comparing random sets of latent vectors of images. Each random set is be shifted by our trained models and by the feature axis regression method (we have mentioned in \ref{sec:feature_axis}) as our baseline. We have generated images from the shifted and not shifted vectors and evaluate them. We are looking for the desired and undesired manifested features.

To add a single feature, we generate the shifted latent vectors by using label vectors of ones for our model and for the baseline approach by shifting the latent vector in the direction of the feature axis with one multiplier. After generating shifted latent vectors, we generate corresponding images using a StyleGAN3 generator with a truncation PSI 0.7. We generate five rows per figure for each of the eleven images, each representing images from the original latent vector, a latent vector shifted by the baseline approach, and a latent vector shifted by our latent feature shifter neural network approach.  

\begin{figure}
    \begin{centering}
        \includegraphics[width=\columnwidth]{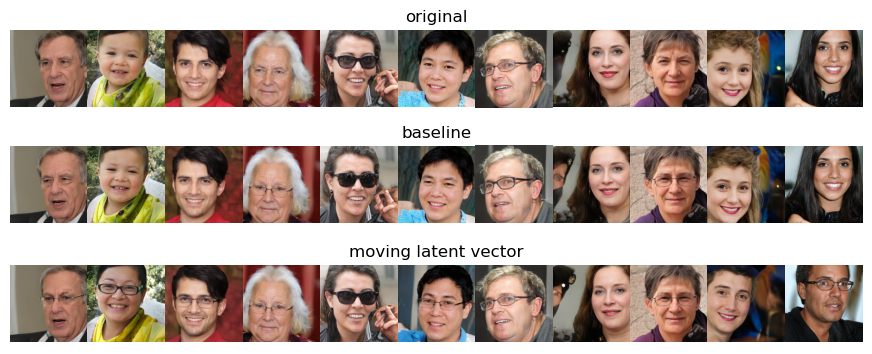}
        \caption{Results of adding eyeglasses feature to eleven random vectors. Each row represents a different approach to adding the feature.\label{fig:eva_eyeglasses}}
    \end{centering}
\end{figure}

We have evaluated our approach by adding the eyeglasses feature at first. In figure~\ref{fig:eva_eyeglasses} we can see that our latent feature shifting approach has the most images with eyeglasses on it. We also noticed that adding eyeglasses to the second image modified with our latent feature shifting approach also made the person look older. This phenomenon is most likely defined by the characteristics of the latent space of StyleGAN3, which was trained on the FFHQ dataset, that probably contains very few (or none) images of infants with eyeglasses and therefore shifting the latent vector into eyeglasses space also made it to move into a negative direction of the "young" feature. 

We have also noticed in Fig.~\ref{fig:eva_eyeglasses} that the last image modified with our latent feature shifting approach contains a man instead of a woman. We suspect that our shifted latent vectors dataset (or even the FFHQ dataset) was imbalanced in a way that the majority of generated persons wearing eyeglasses were male. We can also see that for our latent feature shifting approach (but also the baseline approach) it is more difficult to add eyeglasses to a woman than to a man.

\begin{figure}
    \begin{centering}
        \includegraphics[width=\columnwidth]{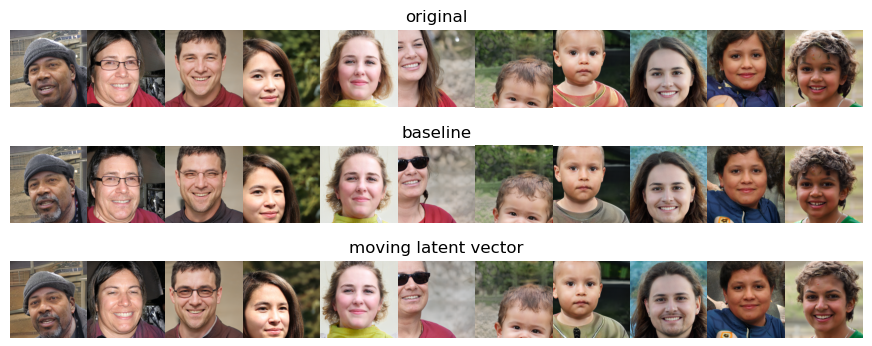}
        \caption{Results of adding the male feature to eleven random vectors. Each row represents a different approach to adding the feature.\label{fig:eva_male}}
    \end{centering}
\end{figure}

As a second evaluation, we have added a male feature to another random set of images to measure the effectiveness. We can see from Figure \ref{fig:eva_male} that our latent feature shifting approach was unsuccessful except for images six and nine, where the baseline approach was not able to add the male feature. To determine whether we have reach an improved compared to the baseline approach, a more extensive evaluation had to follow.

\begin{figure}
    \begin{centering}
        \includegraphics[width=\columnwidth]{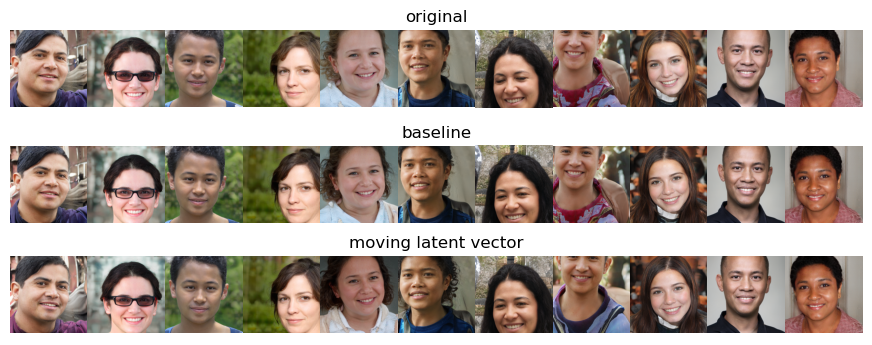}
        \caption{Results of adding black hair feature to eleven random vectors. Each row represents a different approach to adding the feature.\label{fig:eva_bh}}
    \end{centering}
\end{figure}

The last single feature that we evaluated was changing the person's hair to black (black hair feature). Looking at the generated images in Figure \ref{fig:eva_bh}, we can see that our latent feature shifting approach was, in some cases, a little better than the baseline approach when we look at images five and nine. Adding black hair with this approach also added some unwanted hair growth and clothes changes in the resulting image.

We have generated random latent vectors similar to the single feature evaluation to add multiple features with our latent feature shifter neural network model approach. The difference is that for each feature, we have the same model architecture (\textit{a}) but trained on a shifted latent vectors dataset for each feature separately, ending up with three models in total.

\begin{figure*}
    \begin{centering}
        \includegraphics[width=0.7\textwidth]{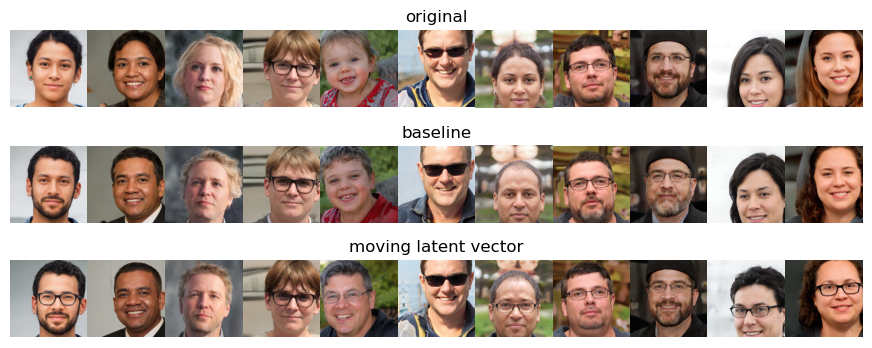}
        \caption{Results of adding eyeglasses and male features to eleven random vectors. Each row represents a different approach to adding the feature.\label{fig:eva_eyeglasses_male}}
    \end{centering}
\end{figure*}

Figure \ref{fig:eva_eyeglasses_male} evaluates adding two features - eyeglasses and male. We can see that at least one feature manifested in most images was modified by our shifting latent vector method. We should also mention a property better in our latent feature shifting approach than the feature axis regression approach. This property can be seen in image number five. Our shifting latent vector method did not further move the vector (which would probably cause unwanted features to manifest, like cloth color change) but instead remained in the same position.   

\begin{figure}
    \begin{centering}
        \includegraphics[width=\columnwidth]{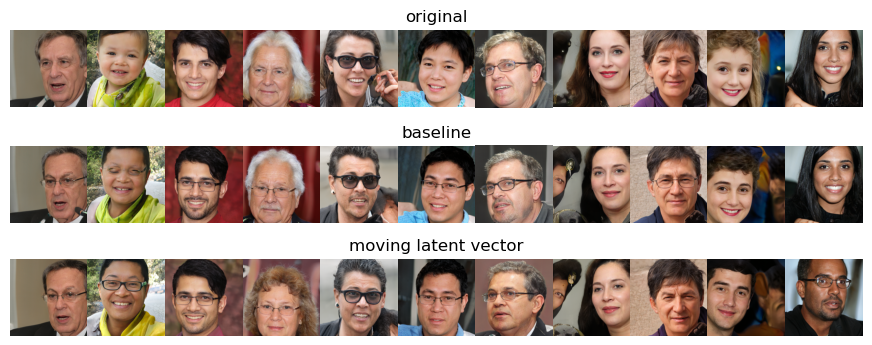}
        \caption{Results of adding eyeglasses male and black hair feature to eleven random vectors. Each row represents a different approach to adding the feature.\label{fig:eva_eyeglasses_male_bh}}
    \end{centering}
\end{figure}

Our last visual evaluation, shown in Figure \ref{fig:eva_eyeglasses_male_bh}, was adding all three features - eyeglasses, male and black hair. In most images modified by our shifted latent vector approach, at least one of the features emerged. The single image was almost unmodified by the baseline and our approach was image number eight.

\section{Evaluation using ResNet34 classifier \label{sec:eva_resnet}}
The manual visual evaluation does not give us much information about how much better our approach was compared to the baseline approach, mainly because it was done on a small set of images. To evaluate more thoroughly, we generated a set of 1000 random images. We have shifted the latent vectors of these images with baseline and latent feature shifting approach as we did in section \ref{sec:eva_manual}. After that, we generated images for each set of shifted images and classified them with the ResNet34 classifier described in section \ref{sec:des_classifier}. We have processed the classifier predictions with a threshold of 0.5 and counted the number of ones (as in terms of standard accuracy evaluation). The important thing to remember while evaluating the ResNet34 classifier is that it was trained with a multi-label classification error of 93\%. Therefore, this error will be transferred to our evaluation results.

We have started by adding a single feature. We can see in Table~\ref{tab:eva_resnet_single} that when adding the eyeglasses feature, our approach outperformed the baseline approach. For the eyeglasses feature, our latent feature shifting (LFS) neural network increased the success ratio of persons being properly classified as wearing eyeglasses from 8.6\% to 54\%, compared to our baseline, which was able to increase it to 30.6\%. 

When adding the male feature, our LFS approach performed similarly to the baseline approach. Each approach increased the number of generated male persons from 63.9\% to 89\%. We can also see that the number of eyeglasses was increasing, but the number of people with black hair was decreasing. We can see that fewer unwanted changes emerged when adding a male feature with our LFS approach compared to the baseline approach.

The last feature evaluated using the ResNet34 classifier is the black hair feature. In adding this feature, our LFS approach did not perform as well as in adding other features, but we can see that our LFS approach has still successfully outperformed the baseline approach. The LFS approach increased the percentage of generated persons with black hair from 11.5\% to 29.4\%, and the baseline approach increased it to 28.4\%. 

\begin{table*}
    \caption{Count of images with added eyeglasses, male and black hair feature that were classified by ResNet34 classifier while adding a single feature. Each row represents a different feature count, and each column represents a different approach.}
    \label{tab:eva_resnet_single}
    \begin{adjustbox}{width=0.7\textwidth, center}
        \begin{tabular}{lccccccc}
        \multicolumn{8}{c}{Evaluation using ResNet34 - single feature (1000 samples)}                                                                                                                            \\ \hline
        \multicolumn{1}{l|}{Added feature}    & \multicolumn{1}{c|}{-}        & \multicolumn{2}{c|}{Eyeglases}               & \multicolumn{2}{c|}{Male}                        & \multicolumn{2}{c}{Black hair} \\ \cline{2-8} 
        \multicolumn{1}{l|}{Approach}         & \multicolumn{1}{c|}{Original} & Baseline & \multicolumn{1}{c|}{LFS}          & Baseline     & \multicolumn{1}{c|}{LFS}          & Baseline     & LFS             \\ \cline{2-8} 
        \multicolumn{1}{l|}{Eyeglasses count} & \multicolumn{1}{c|}{86}       & 306      & \multicolumn{1}{c|}{\textbf{540}} & 159          & \multicolumn{1}{c|}{117}          & 78           & 74              \\
        \multicolumn{1}{l|}{Male count}       & \multicolumn{1}{c|}{639}      & 766      & \multicolumn{1}{c|}{836}          & \textbf{890} & \multicolumn{1}{c|}{\textbf{890}} & 611          & 623             \\
        \multicolumn{1}{l|}{Black hair count} & \multicolumn{1}{c|}{115}      & 94       & \multicolumn{1}{c|}{78}           & 81           & \multicolumn{1}{c|}{101}          & 284          & \textbf{294}    \\ \hline
        \end{tabular}
    \end{adjustbox}
\end{table*}

The second part of the ResNet34 evaluation considered adding multiple features simultaneously. We have chosen to add eyeglasses together with the male feature and all three features together. When looking at the results from Table \ref{tab:eva_resnet_multi}, we can see that our LFS approach was able to increase significantly the number of images containing a person wearing eyeglasses (from 8.6\% to 59.6\%) and the number of images having a male person (from 63.9\% to 96.9\%) when combining these two features simultaneously compared to adding them separately.

When we added all three features simultaneously, we had the best results with the LFS approach compared to the baseline approach. Eyeglasses feature count increased from 8.6\% to 54.5\%, male feature count increased from 63.9\% to 95.6\%, and the black hair feature count increased from 11.5\% to 21.3\%. Eyeglasses and male feature counts were also more significant in the LFS multiple-feature approach than in the LFS single-feature approach.

\begin{table*}
    \caption{Count of images with added eyeglasses, male and black hair feature that were classified by ResNet34 classifier while adding multiple features. Each row represents a different feature count, and each column represents a different approach.}
    \label{tab:eva_resnet_multi}
    \begin{adjustbox}{width=0.7\textwidth, center}
        \begin{tabular}{lccccc}
        \multicolumn{6}{c}{Evaluation using ResNet34 - multiple features (1000 samples)}                                                                                  \\ \hline
        \multicolumn{1}{l|}{Added feature}    & \multicolumn{1}{c|}{-}        & \multicolumn{2}{c|}{Eyeglases + Male} & \multicolumn{2}{c}{Eyeglases + Male + Black hair} \\ \cline{2-6} 
        \multicolumn{1}{l|}{Approach}         & \multicolumn{1}{c|}{Original} & Baseline  & \multicolumn{1}{c|}{LFS}  & Baseline      & LFS                               \\ \cline{2-6} 
        \multicolumn{1}{l|}{Eyeglasses count} & \multicolumn{1}{c|}{86}       & 399       & \multicolumn{1}{c|}{\textbf{596}}  & 380           & \textbf{545}     \\
        \multicolumn{1}{l|}{Male count}       & \multicolumn{1}{c|}{639}      & 941       & \multicolumn{1}{c|}{\textbf{969}}  & 942           & \textbf{956}     \\
        \multicolumn{1}{l|}{Black hair count} & \multicolumn{1}{c|}{115}      & 63        & \multicolumn{1}{c|}{51}   & 200           & \textbf{213}     \\ \hline
        \end{tabular}
    \end{adjustbox}
\end{table*}

\section{Conclusion}
We have implemented and evaluated our novel approach to the controlled generation of generative adversarial networks. We have chosen to demonstrate and evaluate our methods on the StyleGAN3 model, one of the current state-of-the-art models in realistic image generation. We aimed to control semantic facial features on the images generated by the StyleGAN3 generator.

Based on the evaluation we conclude that we have successfully designed the latent feature shifting approach to controlling the generation of the StyleGAN3 generator. This approach may be used in the real-time generation and semantic feature editing of realistic images of human faces. We can also see that the desired features did not manifest in some cases, which creates room for improvement in our approach. Nevertheless, we think that this method, with further improvements, can be used to create and manipulate images similar to police facial composites.

For future work, we would like to improve our LFS approach by creating a better shifted latent vectors dataset to train it with a loss function that would penalize feature entanglement. A human-based classification would be the best solution to ensure we have the image pairs with the right features. When improving the shifted latent vectors dataset, we should balance the images containing certain features to minimize feature entanglement. 

\bibliographystyle{unsrt}
\bibliography{references}

\end{document}